# Research on Personalized Medical Intervention Strategy Generation System based on Group Relative Policy Optimization and Time-Series Data Fusion

Dingxin Lu*

Icahn School of Medicine at Mount Sinai, New York, NY, USA, sydneylu1998@gmail.com

Shurui Wu

Weill Cornell Medicine, New York City, NY, USA, shuruiwu215@gmail.com

Xinyi Huang

University of Chicago, Chicago, IL, USA, bellaxinyihuang@gmail.com

**Abstract**

With the timely formation of personalized intervention plans based on high-dimensional heterogeneous time series information has become an important challenge in the medical field today. As electronic medical records, wearables and other multi-source medical data are increasingly generated and diversified. In this work, we develop a system to generate personalized medical intervention strategies based on Group Relative Policy Optimization (GRPO) and Time-Series Data Fusion: First by incorporating relative policy constraints among the groups during policy gradient updates adaptive balance the individual gain and group gain distribution. To improve the robustness and interpretability of decision-making, the multi-layer neural network structure was employed to group code the patient characteristics. Secondly, for the rapid multi-modal fusion of multi-source heterogeneous time series, a multi-channel neural network combined with self-attention mechanism was employed for dynamic feature extraction, the key feature screening and aggregation were further achieved through the differentiable gating network. Finally, a collaborative search process was proposed to find the ideal candidate intervention strategy based on the combination of genetic algorithm and Monte Carlo tree search so that a global optimization of the candidate intervention strategies was achieved, which greatly enhanced the accuracy of the system as well as the system's personalization level. The experimental results show that model achieves significant improvement in aspects of accuracy, coverage and decision-making benefits of intervention effect compared with existing methods.

CCS CONCEPTS

Applied computing ~ Life and medical sciences ~ Health care information systems

## 1 Introduction

Modern medicine needs personalized medical intervention strategies that can not only enhance patients' treatment effect but also maximize the use of medical resources. In traditional medicine, practitioners frequently apply generic treatment protocols that can address the needs of most patients but fail to consider unique individual differences. Personalized medicine is the science of customizing the medical treatment of individual patients according to their genetic signature, biological markers, behaviour, and environmental exposure. By utilizing this personalized data, precision medicine is able to not only maximise diagnosis accuracy, but actually optimise treatment strategies and diminish the likelihood for adverse reactions, thus ameliorating the overall quality of medicine [1]. In an era of increasingly data-driven healthcare, the generation of personalized strategies for medical intervention has become a critical area of focus for contemporary medical research and clinical implementation.

These challenges are compounded by the diversity and complexity of healthcare data, which can hinder the uptake of personalized medicine strategies. In particular, how to fuse multi-source heterogeneous data, and, as a key issue to achieve precision medicine, how to lay around the time series data processing. Medical data includes structured data such as electronic health records, unstructured data such as clinical notes, genetic sequencing data, and wearable device sensor data, as well as medical imaging data [2]. Time-series data – like continuous physiotherapy monitoring data, heart rate variability, blood glucose levels, and electroencephalogram signals, etc. – is one of them and is crucial in monitoring the variations between patient health states. But through time series data is complex, high-dimensional, dynamic change characteristics, which make it difficult for traditional data analysis methods to effectively mine the key information and give accurate medical intervention suggestions [3]

Time-series data is ubiquitous in current medical practice such as continuous physiological signals from wearable health equipment, regular laboratory tests and time-based health history of patients. Collectively, these data sources offer dynamically evolving information on patient health and empower the physician to perform abnormal detection at an early stage, predict disease progression, and modify personalized treatment over time [4]. But time series data is typically high dimensional, very noisy, and asynchronous, posing great challenges for both traditional statistics models and machine learning approaches. Time series data often has intricate dependencies and non-linear features that are difficult to extract using traditional linear regression or rule-based analysis techniques. Moreover, issues like incomplete data, data sparsity, and sensor malfunction further complicate data interpretation, thereby impacting the accuracy of personalized medical choices [5].

The development of techniques for efficiently processing and fusing time-series data is a key enabler of the generation of actionable, accurate personalized medicine strategies. The past few years have witnessed a surge of developments in artificial intelligence (AI) technologies, particularly deep learning and reinforcement learning [13, 17-19], which are now widely used for the analysis of large-scale medical and social data [16,22]. In particular, they are well suited for time-series data which requires capturing long-term dependencies such as with recurrent neural networks (RNNs) and some of their variants (LSTM and GRU) that has been used in ECG signal extraction and arrhythmia detection tasks. Moreover, the attention mechanism founded on the Transformer architecture could address data from a longer time series scope with higher efficiency and identify complex feature interactions, which shows great potential to analyze medical time series data.

## 2 RELATED WORK

Initially, Motevall et al. [6] and Wu and Huang [16] addressed the shift from broad, generalized intervention strategies to personalized ones, especially among low-income populations based on structured-unstructured EHR data fusion. Research has already highlighted that current obesity management approaches are largely one-size-fits-all and do not consider individuals' heterogeneous characteristics. Personalized medicine, however, is able to formulate more precise treatment approaches for every patient, using data on the child's genetic aspects, lifestyle, dietary requirements, and other factors. Obesity highlights the uniqueness of individuals and the impact of their interactions—they're not just numbers; composed of factors including, nutrition, education, and exercise that are essential to develop tailored interventions to combat obesity.

Additionally, Santaló et al. [7] deepen knowledge of the ethical implications of epigenetics in personalized medicine. Thanks to developments in genomics and epigenetics, personalized medicine is starting to better target treatment based on a person's genetic makeup and environmental exposures. Furthermore, this individualized therapy has also sparked some ethical debates, particularly as the employment of a person's genetic information may accentuate social inequalities or affect public health policy.

Furthermore, Gupta et al. [8] emphasized that the broad application of AI technology in big data management and analysis will become more important with the rapid expansion of medical data. AI enables patients to get more accurate treatment options by quickly processing and reviewing data from various medical sources. AI can assist medical professionals to formulate personalized treatment strategies as well as automate diagnosis and treatment, reduce human error, and improve patient outcome. Parekh et al. [9] underlying this is a deep learning AI model that predicts personalized medicine choices on the basis of the genetic background and patient history. The technology can assist healthcare providers to customize treatment plans based on a patient's genomic data and medical history, including drug treatment and disease prevention, the researchers said. The use of AI algorithms that analyze extensive datasets of genetic markers and health records means identifying the most critical determinants of any specific patient health and designing more targeted treatment regimens.

Additionally, Bonkhoff et al. [10] explored the utilization of artificial intelligence in estimating individual treatment outcomes. Because treatment and rehabilitation processes for stroke can be highly individualized, predicting how an individual will respond to treatment is key to optimizing treatment strategies. With the advent of big data and the development of artificial intelligence technology, more and more patient data can be applied by healthcare providers to establish personalized treatment models, according to research. These models not only allow doctors to determine effectiveness of different treatment options but also give patients a personalized plan.

Collin et al. [11] recommendations for data integration and model validation based on computational models in personalized medicine Computational models are becoming increasingly important in patient stratification, disease prediction, and treatment effect evaluation alongside the evolving personalized medicine. These models combine multi-source heterogeneous data (genomic information, clinical data, imaging data, etc.) to help healthcare professionals better understand the pathological characteristics of patients and make personalized treatment plans. Now, a complex ensemble of computational approaches to personalized medicine and best practices for data integration and model validation are discussed in detail in work led by Rakesh Kumar.

Vadapalli et al. [12] explained the use of artificial intelligence and machine learning techniques in analysing gene expression and variant data to enhance personalised medicine. Gene expression data and variant information in precision medicine are becoming more prominent with the rapid advance in genomics. AI/ML methods have been utilized to analyze vast amounts of available genetic data to discover variants that drive disease as well as tailor treatment strategies in accordance with the genomic properties of individuals. These 32 AI/ML methods cover a range of types, including classification models, regression analysis, clustering algorithms, etc. This study discusses these applications in a systematic manner and adds to the body of literature demonstrating how efficiently these methods can be used in genomics and precision medicine.

## 3 METHODOLOGIES

### 3.1 Group Relative Policy Optimization

We first model the personalized medicine intervention task as a Markov decision process (MDP), so that for the patient set $\{1,2,\ldots,N\}$, each patient $i$ has a state $s_t^{(i)}$ at any time $t$ and the executable intervention action is $a_t^{(i)}$. We want to learn a strategy $\pi_\theta(a|s,i)$ to maximize long-term cumulative rewards. The multimodal feature of patient $i$ is denoted as $x_i$, through a multi-layer network $\Phi(x_i)$ to obtain the patient embedding vector $e_i \in \mathbb{R}^d$. We clustered $\{e_i\}$ to obtain $K$ groups, and each patient had a group tag $g_i \in \{1,2,\ldots,K\}$. In order to take into account the individual optimal and group optimum, we introduce additional penalty terms and group mean terms into the dominance function. Let the $Q$ function and value function of an individual be Equations 1 and 2 respectively:

$$Q_i^\pi(s,a) = \mathbb{E}\left[\sum_{\tau=0}^\infty \gamma^\tau R\left(s_{t+\tau}^{(i)}, a_{t+\tau}^{(i)}\right) \middle| \left(s_t^{(i)}, a_t^{(i)}\right) = (s,a)\right], \quad (1)$$

$$V_i^\pi(s) = \mathbb{E}\left[\sum_{\tau=0}^\infty \gamma^\tau R\left(s_{t+\tau}^{(i)}, a_{t+\tau}^{(i)}\right) \middle| \left(s_t^{(i)}\right) = (s)\right]. \quad (2)$$

Then the individual dominance function is Equation 3:

$$A_i^\pi(s,a) = Q_i^\pi(s,a) - V_i^\pi(s). \quad (3)$$

For patients in group $g$, the set $G_g = \{j|g_j = g\}$, which defines the population mean $Q$ function and the value function as Equations 4 and 5:

$$Q_g^\pi(s,a) = \frac{1}{|G_g|} \sum_{j \in G_g} Q_j^\pi(s,a), \quad (4)$$

$$V_g^\pi(s) = \frac{1}{|G_g|}\sum_{j \in G_g} V_j^\pi(s). \quad (5)$$

This results in the average dominance function of the group, as shown in Equation 6:

$$A_g^\pi(s,a) = Q_g^\pi(s,a) - V_g^\pi(s). \quad (6)$$

On this basis, we define the relative advantage of the group $\tilde{A}_i^\pi(s,a)$. In order to make it more flexible and punitive, we further break it down into three parts (which can be combined or simplified as needed during implementation), such as Equation 7:

$$\tilde{A}_i^\pi(s,a) = \alpha_1 A_i^\pi(s,a) + \alpha_2 A_{g_i}^\pi(s,a) - \alpha_3 \parallel A_i^\pi(s,a) - A_{g_i}^\pi(s,a) \parallel^\beta, \quad (7)$$

where $\alpha_1$, $\alpha_2$ and $\alpha_3$ are the hyperparameters used to balance individual gains, group gains, and individual-group difference punishments, and $\beta$ can take 2 or other positive numbers to control the sensitivity of punishment. When $\beta = 2$, it is the square norm, which can give a stronger penalty when the difference between the average return of the individual and the group is too large.

The intuitive implication of this is: when $\alpha_1$ is dominant, more emphasis is placed on individual optimality; When $\alpha_2$ is dominant, the group optimum is emphasized. When $\alpha_3$ is large enough, the difference between individual and group returns will be strongly suppressed, so as to ensure a certain degree of "fairness". When updating a policy, refer to the idea of Proximal Policy Optimization (PPO), so that $\theta$ and $\theta_{old}$ represent the parameters of the current policy network and the old policy network respectively. Define the probability ratio as shown in Equation 8:

$$\rho_i(\theta) = \frac{\pi_\theta(a|s,i)}{\pi_{\theta_{old}}(a|s,i)}. \quad (8)$$

The clipping mechanism is used to constrain the update amplitude. The traditional PPO objective function can be written as Equation 9:

$$L_{PPO}(\theta) = \mathbb{E}\left[\min\left(\rho_i(\theta)A_i^\pi(s,a), clip(\rho_i(\theta), 1-\epsilon, 1+\epsilon)\tilde{A}_i^\pi(s,a)\right)\right]. \quad (9)$$

In our Population Relative Strategy Optimization (GRPO), we replace $A_i^\pi$ with $\tilde{A}_i^\pi$ and add an additional distribution smoothing and KL penalty for group labels, as shown in Equation 10:

$$\begin{aligned}L_{GRPO}(\theta) = \mathbb{E}_{(s,a,i)\sim D}\\ \left[\min\left(\rho_i(\theta)\tilde{A}_i^\pi, clip(\rho_i(\theta), 1-\epsilon, 1+\epsilon)\tilde{A}_i^\pi(s,a)\right)\right]\\ -\lambda_{KL}\sum_{g=1}^K KL\left(\pi_{\theta_{old}}^g \parallel \pi_\theta^g\right),\end{aligned} \quad (10)$$

the $\pi_\theta^g$ can be understood as the distribution of the strategies learned under the condition of group $g$ (e.g., fine-tuning the bias of the policy network output layer by group or additional network branches), $KL(\cdot\|\cdot)$ represents the KL divergence, and $\lambda_{KL}$ controls the drastic changes in the distribution of the old and new strategies in each group to ensure the stability of the training process. In this way, the strategy can find a compromise between individual and group benefits among multiple groups, and reduce the training instability caused by large group differences.

In our method, Population Relative Strategy Optimization (GRPO) and the multi-layer neural network structure are tightly coupled. The GRPO balances the merits of individuals and groups in the optimization process, in order to optimize personalized medical intervention strategies. The multi-channel neural network with differentiable gating network can extract high-quality deep features from multi-modal time series data, and enhance system robustness and accuracy.

## 3.2 Feature Extraction and Collaborative Search

After obtaining the population relative strategy optimization framework, we still need to extract high-quality state representations from multi-source time series features to support the accuracy of intervention decisions. To this end, for M data sources of different modalities (such as physiological signals, test indicators, clinical texts, etc.), we use a parallel multi-channel structure and a self-attention mechanism for deep fusion, and add a differentiable gating network to screen key features at the backend. Then, the cooperative search of genetic algorithm and Monte Carlo tree search is used to find a better solution in the high-dimensional action/strategy space.

Let $X^{(m)} \in \mathbb{R}^{T \times d_m}$ represent the sequential input of modal $m$, and the output $H^{(m)} \in \mathbb{R}^{T \times h}$ can be obtained after passing through the $f_{\theta_m}$ of a temporal network (such as LSTM or 1D convolution). In order to capture long-term dependencies and intermodal interactions, we further apply Multi-Head Self-Attention (MHA) to $H^{(m)}$. The single-head self-attention is given in Equation 11:

$$Attention(Q, K, V) = softmax\left(\frac{QK^T}{\sqrt{d_k}}\right)V, \qquad (11)$$

where $Q$, $K$ and $V$ are obtained by the $H^{(m)}$ linear mapping. Multi-head self-attention performs multiple attention in parallel, and splices and linear transformations after output. Remembering that the long attention operation is $MHA(\cdot)$, then there is equation 12:

$$Z^{(m)} = MHA(H^{(m)}). \qquad (12)$$

And stitch together the results of each modality, as in Equation 13:

$$Z = [Z^{(1)} \parallel Z^{(2)} \parallel \cdots \parallel Z^{(M)}] \in \mathbb{R}^{T \times (M \cdot h)}. \qquad (13)$$

To further suppress the noise signature and highlight the information of the critical modal/timing positions, we add a differentiable gating network after $Z$, as shown in Equations 14 and 15:

$$G = \sigma(W_g Z + b_g), \qquad (14)$$

$$F = G \odot Z, \qquad (15)$$

where $W_g$ and $b_g$ are learnable parameters, and $\sigma(\cdot)$ It can be Sigmoid activation, $G$ and $Z$ elements are in the same dimension, and $\odot$ is Hadamard element-by-element multiplication. In this way, we can obtain $F \in \mathbb{R}^{T \times (M \cdot h)}$, which will automatically learn to pay attention to the key modal-temporal position in the training, and filter out irrelevant or noisy information.

We encode a strategy or a sequence of intervention actions as a chromosome, such as $\mathcal{X}$, and define the fitness function as Equation 16:

$$Fitness(\mathcal{X}) = \mathbb{E}\left[\sum_{t=0}^{T} \gamma^t R(s_t, a_t(\mathcal{X}))\right], \qquad (16)$$

Selection, Crossover, and Mutation iterations are used to find a number of highly fit candidate $\{\mathcal{X}_1^*, \mathcal{X}_2^*, \ldots, \mathcal{X}_L^*\}$ in a large parameter/action sequence space. For the above candidates, we further employ Monte Carlo tree search to refine locally. Let $\mathcal{T}$ represent the search tree in the state space, and expand it down with "the root node as the initial state". The MCTS consists of four steps, as shown in Equation 17:

$$Selection \rightarrow Expansion \rightarrow Simulation \rightarrow Backpropagation. \qquad (17)$$

In the selection stage, the upper confidence interval (such as UCB) is used to balance the explored value and unexplored child nodes, and when expanding, the possible actions of new nodes are added and random simulations or policy network sampling are performed to evaluate the returns, and then the number of visits and value accumulation of nodes are updated back to the lake. If $MCTS(\mathcal{X})$ represents the optimal return after several Monte Carlo tree searches on a chromosome (or strategy) $\mathcal{X}$, then in the batch of candidates obtained in the GA iteration, we can finally choose $\mathcal{X}^*$ as follows in Equation 18:

$$\mathcal{X}^* = \underset{\mathcal{X} \in \{\mathcal{X}_1^*, \mathcal{X}_2^*, \ldots, \mathcal{X}_L^*\}}{\mathrm{argmax}} MCTS(\mathcal{X}). \qquad (18)$$

The whole process can be regarded as using genetic algorithm to find the optimal solution domain, complemented by practical optimization strategies commonly seen in real-world scenarios such as advertising systems and credit risk monitoring [14-15].

# 4  Experiments

## 4.1 Experimental Setup

The dataset used in this experiment is MIMIC-III, a large, freely-available database of real-world intensive care unit (ICU) patient data that is utilized across medical data analytics and artificial intelligence research. The MIMIC-III dataset, collected by the Beth Israel Deaconess Medical Center (BIDMC) in Boston, USA, contains extensive amounts of clinical information, including information about the diagnosis and treatment, drugs used, laboratory test results, vital signs, etc. of patients admitted in ICU. The MIMIC-III dataset includes detailed observations of over 60,000 ICU patients who were admitted between 2001 and 2012. The datasets contain rich structured and unstructured information, such as: patient diagnostic codes (ICD-9), drug prescriptions, laboratory data, and vital signs (blood pressure, heart rate, respiratory rate, etc.) as well as medical logs. These data are particularly well suited for studying personalized medical interventions, as they provide a generally broad.

We will explore the MIMIC-III dataset of patient diagnosis, medication records, vital signs and laboratory based on this data through machine learning and analysis techniques to learn to build a predictive model to provide patients Next treatment recommendation. To explore the intelligent medical approach, this paper investigates the MIMIC-III data to provide an intelligent medical decision support system, which can provide a personalized and accurate treatment plan for the clinician based on real-time inspection data (including diagnosis, medication, laboratory data etc.) to improve the effect of treatment and reduce unnecessary medical intervention.

## 4.2 Experimental Analysis

In order to validate the effectiveness of our method, the experiment we conducted will be compared with four kinds of widely-used benchmark methods listed as follows: logistic regression (LR), which uses a linear model to predict classification outcomes, is suitable for simple tasks but is hard to capture complex data relationships; Random forest (RF) is an ensemble learning method that can process high-dimensional data and increase the prediction accuracy, but the model is complicated and the interpretation is difficult; Support vector machine (SVM) is method that maximizes the classification gap to optimize classification accuracy, which is suited to high-dimensional data but with large amount of computing. and deep neural networks (DNNs), which can implicitly extract complex features, learn end to end, and achieve state-of-the-art performance when treating large-scale complex data, but are difficult to train and interpret.

Figure 1 the accuracy of different learning methods during training. The Figure 1 results can observe that the Ours method from the original 0 can continue to increase and ultimately over converges to a higher accuracy with the greater number of training rounds, indicating that the training process is stable and can eventually achieve relatively ideal performance. However, while other existing methods also have gradually improving performances with the training progress, the performances fluctuate greatly and the final accuracy does not converge to a high level as Ours method. The variability of these methods suggest that they are very sensitive to noise during training and, as a result, the final performance is limited.

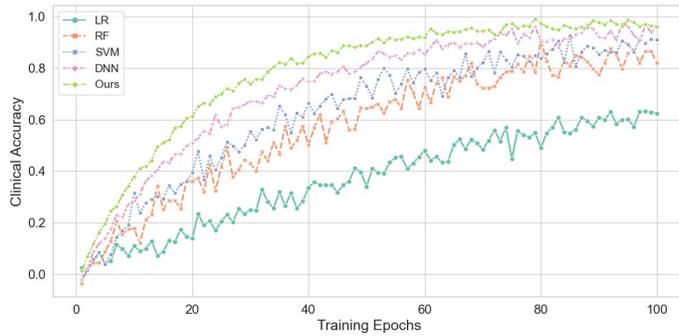

**Figure 1.** Comparison of Clinical Accuracy for Different Methods.

In Figure 2, we present the performance of each model in terms of AUC distribution across all combinations of learning rate and number of rounds. Boxline and density can be combine as plotge in order to visualize the shapes of distribution, median, interquartile ranges and data probability densities graph as plot.

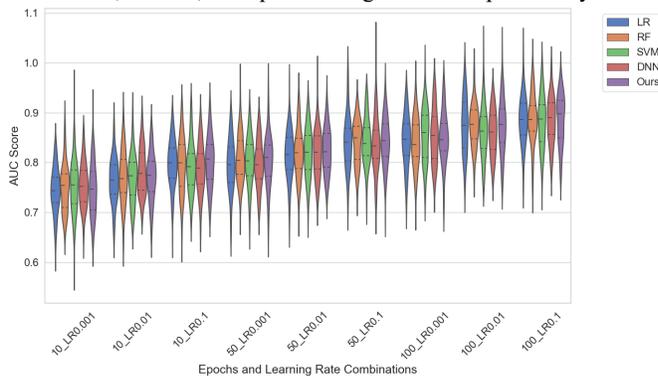

**Figure 2.** Decision-making Benefits: Model Comparison across Different Epochs and Learning Rates.

Figure 2 shows the wide variation in distribution of AUCs achieved by different models settings at the same parameter combinations. Notably, our approach (Ours) achieved greater median AUC and a narrower distribution over most ranges in the hyperparameter space, reflecting both a superior and more robust performance. Whereas the AUC distribution of other models is a wide scattered distribution, and it has a low median, which indicates performance instability.

In the comparative experiment between the emergency medical scenario and the chronic disease management scenario, the Ours method showed excellent real-time response ability and long-term stability. The rapid decision-making in emergency scenarios and the long-term prediction ability in chronic disease management.

# 5 Conclusion

In conclusion, we described a personalized medical intervention strategy generation system based on population relative strategy optimization and time series data combination, and proposed personalized treatment recommendation system based on the MIMIC-III dataset data together with patients diagnosis, medication record, vital signs and laboratory data. The experimental results demonstrate that our approach achieves better Decision-Making benefits than traditional models. In the future, as medical big data and artificial intelligence technology continue to develop—such as LLM applications in mental health [13], misinformation detection [22], and model optimization [19-20]—personalized medicine will become increasingly accurate and adaptive to patient needs. But there are challenges to be resolved: data privacy protection, model interpretability and the feasibility of clinical application. Future research can focus on the fusion of multi-source heterogeneous data, improve the interpretability of models, and solve practical problems such as real-time and data privacy protection.